%% file: main.tex
\documentclass[10pt,twocolumn,letterpaper]{article}

\usepackage[pagenumbers]{cvpr} %

\input{preamble}

\input{macros}

\definecolor{cvprblue}{rgb}{0.21,0.49,0.74}
\usepackage[pagebackref,breaklinks,colorlinks,citecolor=cvprblue]{hyperref}

\def\paperID{15456} %
\def\confName{CVPR}
\def\confYear{2024}

\title{Artist-Friendly Relightable and Animatable Neural Heads}

\author{Yingyan Xu$^{1, 2}$
	\hspace{7mm}
	Prashanth Chandran$^{2}$
	\hspace{7mm}
	Sebastian Weiss$^{2}$
	\\
	Markus Gross$^{1, 2}$
	\hspace{7mm}
	Gaspard Zoss$^{2}$
	\hspace{7mm}
	Derek Bradley$^{2}$
	\\
	$^{1}$ETH Z\"urich
	\hspace{7mm}
	$^{2}$DisneyResearch\textbar Studios \\
	{\tt\small {\{yingyan.xu,grossm\}@inf.ethz.ch}} \\ 
	{\tt\small {\{prashanth.chandran,sebastian.weiss,gaspard.zoss,derek.bradley\}@disneyresearch.com}}
}

\begin{document}
\maketitle
\input{sec/0_abstract}
\input{sec/1_intro}

\input{sec/2_relatedwork}
\input{sec/3_method}

\input{sec/4_experiments}

\input{sec/5_conclusion}
{
    \small
    \bibliographystyle{ieeenat_fullname}
    \bibliography{main}
}

\input{sec/X_suppl}

\end{document}

%% file: preamble.tex
\usepackage{microtype}
\usepackage{booktabs}
\usepackage{multirow}
\usepackage{array}
\usepackage{adjustbox}
\usepackage{stfloats}

\usepackage[dvipsnames]{xcolor}


%% file: macros.tex
\newcommand{\figref}[1]{Fig.~\ref{#1}}
\newcommand{\tabref}[1]{Table~\ref{#1}}

\newcommand{\secref}[1]{Section~\ref{#1}}

\newcommand{\shortcite}[1]{\cite{#1}}

\definecolor{dbcolor}{RGB}{50,10,210}

\definecolor{piccolo}{RGB}{10,150,100}

\definecolor{swcolor}{RGB}{210,10,210}

\definecolor{gzcolor}{RGB}{210,210,10}

\definecolor{yxcolor}{RGB}{10,210,50}

\definecolor{delcolor}{RGB}{210,0,0}
\definecolor{addcolor}{RGB}{0,0,0}

\makeatletter
\newcommand\footnoteref[1]{\protected@xdef\@thefnmark{\ref{#1}}\@footnotemark}
\makeatother

%% file: sec/0_abstract.tex
\begin{abstract}
An increasingly common approach for creating photo-realistic digital avatars is through the use of volumetric neural fields.  The original neural radiance field (NeRF) allowed for impressive novel view synthesis of static heads when trained on a set of multi-view images, and follow up methods showed that these neural representations can be extended to dynamic avatars.  Recently, new variants also surpassed the usual drawback of baked-in illumination in neural representations, showing that static neural avatars can be relit in any environment.  In this work we simultaneously tackle both the motion and illumination problem, proposing a new method for relightable and animatable neural heads.  Our method builds on a proven dynamic avatar approach based on a mixture of volumetric primitives, combined with a recently-proposed lightweight hardware setup for relightable neural fields, and includes a novel architecture that allows relighting dynamic neural avatars performing unseen expressions in any environment, even with nearfield illumination and viewpoints.
 
\end{abstract}

%% file: sec/1_intro.tex
\section{Introduction}
\label{sec:intro}

Creating realistic digital avatars of real people has many applications, for example in video games, films, VR experiences and telepresence.  Original methods involved scanning and tracking geometry and appearance properties from one or more cameras and then applying a traditional graphics rendering pipeline to generate novel images.  The challenge lies in the fact that avatars consist of several complex components like skin, eyes, teeth, and hair, each with complex material properties that are difficult to acquire and render realistically. As such, there has been a recent push towards {\em neural} representations of avatars, which forego triangles and texture maps and bypass traditional ray-tracers in exchange for neural rendering, where the avatar is represented by a neural network that can be queried at render-time, with equal handling of skin, eyes, teeth and hair in a single model.

The original Neural Radiance Field (NeRF)~\cite{mildenhall2020nerf} representation laid the ground work for creating neural avatars that could be re-rendered photo-realistically from novel viewpoints.  NeRFs are trained on large collections of images and represent a static scene as an MLP that is queried multiple times during volume rendering. Since the advent of NeRFs, several extended representations have emerged, which aim to increase the rendering speed~\cite{mueller2022instant}, create NeRF avatars from smart phones~\cite{park2021nerfies}, morph between neural avatars~\cite{wang2022morf}, or create generative neural heads~\cite{chan2022eg3d}.

An important component for the adoption of neural avatars is the ability to represent motion, which is not possible with the original neural field approaches.  To account for this, researchers have devised extended neural representations like Neural Volumes~\cite{lombardi2019neuralvolumes}, NeRFBlendShapes~\cite{gao2022nerfblendshape}, NeRSemble~\cite{kirschstein2023nersemble} and a Mixture of Volumetric Primitives (MVP)~\cite{lombardi2021mvp}, which aim to learn dynamic representations of human heads from video data.

A second important aspect of digital avatars is the ability to relight the head in any novel illumination.  As with addressing the motion requirement, dedicated architectures have been proposed to address the relightability requirement, such as NeLFs~\cite{sun2021nelf}, NRTFs~\cite{lyu2022nrtf} and ReNeRF~\cite{xu2023renerf}, which can reliably relight neural representations of static scenes.

In order to have the most flexible and artist-friendly neural head avatars they should be both animatable {\em and} relightable.  In this work we propose a new architecture for neural head avatars that can be relit to match any distant environment map or nearfield light sources, while at the same time providing full dynamic control over the facial shape.  This allows both the playback of captured performances as well as the generation of novel artistically-created performances (e.g. driven by rig controls, retargeting or input video tracking), all with full control over the scene illumination and viewpoint.

Our approach is to build on a dynamic avatar representation that uses a Mixture of Volumetric Primitives (MVP)~\cite{lombardi2021mvp}, which achieves efficient rendering of animatable neural heads with high visual quality.  The idea of MVP is to decode the geometry and appearance of a person-specific facial expression into a collection of geometric primitives containing color and opacity information, which can be sampled during traditional volumetric rendering.  In our work, we propose to condition the appearance branch not just on the view direction but also on the scene lighting, in order to achieve controllable appearance changes based on different lighting conditions at run time.  To achieve this we propose to train the model on captured dynamic performances under one-light-at-a-time (OLAT) illumination, which simplifies the lighting representation to a single light direction vector per frame.  Importantly, we propose to compute per-primitive local light and view directions during the conditioning of the appearance branch, which allows us to represent both nearfield lights and distant environment map illumination, as well as nearfield viewpoints with varying focal length.  We show that our method achieves high quality animatable and relightable neural avatars without the need for training data from a dense light stage, but instead using a less expensive sparse array of LED bars often used in photogrammetry setups.  Once trained, the result is an artist-friendly neural representation of a complete dynamic head that can be controlled via traditional mesh deformation and scene illumination like in the familiar graphics pipeline.

%% file: sec/2_relatedwork.tex
\section{Related Work}
\label{sec:relatedwork}

Realistic digital double creation requires accurate modeling of the face geometry, appearance and motion. In this section, we focus our discussion on image-based relighting, mesh-based representations and more recent work on neural volumetric avatars.

Image-based relighting techniques exploit the linearity of light transport to synthesize the scene under novel illumination conditions as a linear combination of a set of one-light-at-a-time (OLAT) images. Debevec \etal \cite{debevec2000acquiring} were the first to use a light stage to acquire a dense reflectance field of a human face. Sun \etal \cite{sun2019single} proposed a neural network trained on light stage OLAT data that takes as input a single portrait image and directly predicts a relit image given an environment map. Later work \cite{sun2020light} has also explored ways to supersample the fixed basis used during capture to allow continuous high frequency relighting. These methods can only be applied to a static expression as the subject has to stay still during the capturing of OLAT images, and relighting is limited to the captured view point.

Compared to image-based techniques, mesh-based representations can handle novel views by design and offer explicit control of motion if temporally consistent tracking is provided~\cite{3drecsota}. To enable photorealistic rendering of digital humans, reflectance acquisition is also important in addition to geometry. Traditional facial appearance capture systems \cite{ma2007rapid, ghosh2011multiview, gotardo2018practical, riviere2020single} obtain parameters of predefined BRDFs as texture maps via inverse rendering, but are often limited to the skin regions. 
In recent years, mesh-based \cite{thies2019deferred,Riegler2020-bi,Jena2023-ay} or point-based \cite{Aliev2020-lv} neural rendering approaches have gained popularity because they do not assume simplified reflectance models and can deal with imperfect geometry. 
Zhang \etal \cite{zhang2021neural} proposed to model non-diffuse and global illumination as residuals added to a physically-based diffuse base rendering in texture space. But they have shown free viewpoint relighting of only a static expression. 
Meka \etal \cite{Meka:2020} used spherical gradient illumination to allow dynamic capture. However, their method can only be applied to performance playback due to the lack of correspondence between frames. 
Bi \etal \cite{bi2021deep} proposed a deep relightable appearance model (DRAM) as a VAE that takes as input a track mesh and an average texture and outputs the mesh vertices and view-dependent OLAT textures. 
All of these methods share the disadvantages of mesh-based representations, such as thin structures (\eg, hair), semi-transparent shiny materials (\eg, eyes), and large occlusions (\eg, teeth and tongues), which can be difficult to be tracked, reconstructed, or rendered.

More recently, there has been a rise of interest in volumetric representations since Neural Radiance Fields (NeRFs) \cite{mildenhall2020nerf,sitzmann2019srns} were proposed. NeRFs represent the scene using a coordinate-based network that outputs color and density for each point observed from any view direction, trained on multi-view image input along with camera parameters.  However, the original NeRFs are limited to static scenes under a fixed lighting condition. Therefore, motions or scene lighting cannot be modeled or controlled.

Followup work has enabled NeRFs to perform relighting \cite{zhang2021nerfactor, sun2021nelf, lyu2022nrtf, xu2023renerf, toschi2023relight, zeng2023relighting, boss2021neural}. ReNeRFs \cite{xu2023renerf} take the idea of image-based relighting and extend NeRFs with an OLAT MLP and a spherical codebook to allow smooth lighting interpolation without a dense light stage. However, ReNeRFs can only handle static scenes. %

Many methods have extended NeRFs to dynamic scenes. Some commonly used schemes are: (1) use a deformation field for motion and model appearance in a canonical space \cite{li2021neural, park2021nerfies, pumarola2020d,nerfplayer,Tretschk2023-uu}; (2) learn a time-conditioned radiance field \cite{li2022neural, park2021hypernerf}; (3) learn a radiance field for each time step often with spatial decomposition \cite{kplanes_2023, fang2022fast, lombardi2019neuralvolumes, Jang2022-fd, shao2023tensor4d}. While being generic to model any dynamic scenes, these methods are restricted to performance playback, or only with limited motion manipulation capability without semantic control. Various methods \cite{Gafni_2021_CVPR, gao2022reconstructing, zielonka2022instant} have also explored combining 3D morphable models, \eg, FLAME \cite{FLAME:SiggraphAsia2017} with NeRFs, or include skeletal animations~\cite{Habermann2023-jw,Weng_2022_CVPR}.  However, these methods have only limited fidelity for re-animation and do not support relighting.

Different from explicit mesh representations or fully volumetric representations like NeRFs, the Mixture of Volumetric Primitives (MVP) representation \cite{lombardi2021mvp} is a hybrid representation that inherits the strengths of both. MVP models the scene as a collection of geometric primitives with spatially varying color and opacity driven by a coarse guide mesh.  Follow-up work has extended MVP to articulated human bodies \cite{remelli2022drivable}. Iwase \etal \cite{iwase2023relightablehands} combined MVP and the student-teacher relighting framework in DRAM \cite{bi2021deep} and demonstrated the result on articulated hand models. Concurrent with our work, TRAvatar \cite{yang2023towards} extends MVP with a linear lighting branch designed to explicitly follow the linear nature of lighting.  However, this method assumes a fixed basis, which often needs to be very dense for high fidelity relighting, and cannot interpolate/extrapolate novel lighting directions or model nearfield effects.  Our work addresses all of these shortcomings and does not require an expensive light stage.

%% file: sec/3_method.tex
\begin{figure*}[t]
    \centering
    \includegraphics[width=\textwidth]{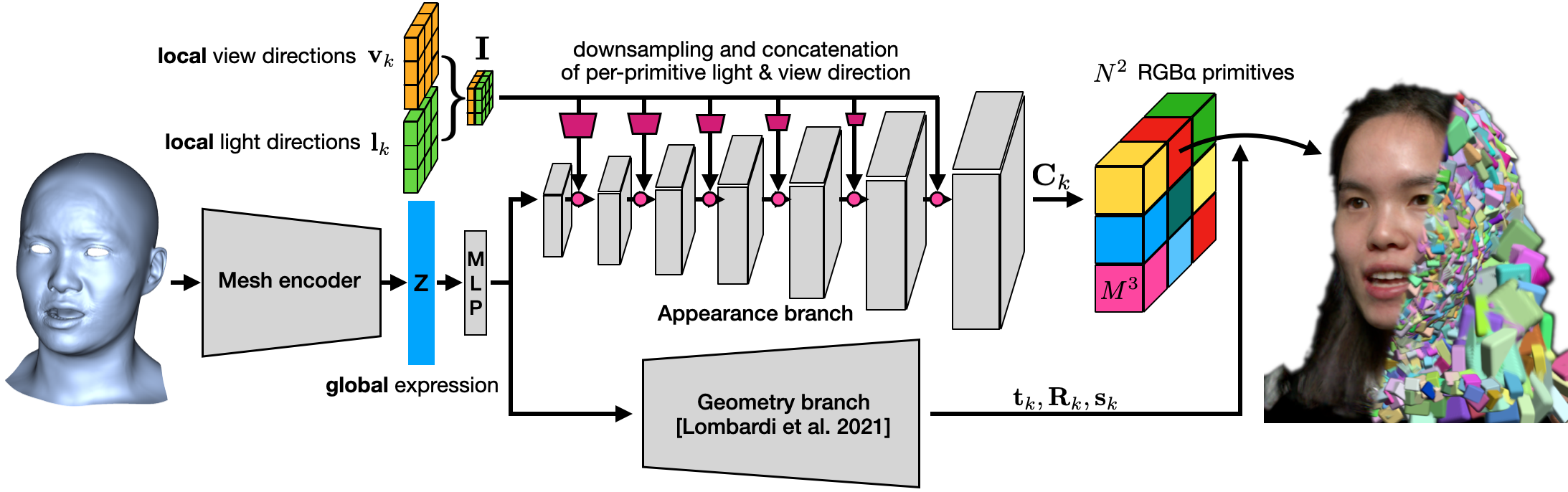}
    \caption{Overview of our pipeline, based on the Mixture of Volumetric Primitives (MVP) architecture~\cite{lombardi2021mvp}. A global expression code $\mathbf{z}$ obtained from a mesh encoder is fed into appearance and geometry decoder branches. The convolutional appearance branch predicts color and opacity of $N^2$ primitives with a grid resolution of $M^3$ each, which are placed in the scene as 3D primitives using the geometry branch (unchanged from Lombardi et al.~\cite{lombardi2021mvp}).  To support relighting and especially near-field lighting, we augment the appearance branch and introduce per-primitive local view and light directions $\mathbf{v}_k, \mathbf{l}_k$ that are concatenated at every layer of the appearance branch network.}
    \label{fig:overview}
\end{figure*}

\section{Relightable and Animatable Neural Heads}
\label{sec:method}

We now describe our method to create relightable and animatable neural heads.  Our approach is to start with the baseline animatable head model of Lombardi et al.~\shortcite{lombardi2021mvp}, which uses a Mixture of Volumetric Primitives (MVP) to describe a person-specific deformable neural head model, and extend the architecture and training data to allow for relighting.  An overview of our method is given in \figref{fig:overview}, and the details are described in the following sub-sections.  We first provide a brief overview of MVP for background information (\secref{subsec:MVP}), and then describe our extensions starting with the data requirements (\secref{subsec:data}), the main architecture (\secref{subsec:relightableMVP}), and implementation details (\secref{subsec:implementation}).

\subsection{Mixture of Volumetric Primitives}
\label{subsec:MVP}

MVP is a state-of-the-art neural representation for human heads~\cite{lombardi2021mvp}.  The key idea is that a collection of simple geometric primitives can be used collectively to render the complex geometry of a human face with high fidelity.  The inputs to MVP include a person-specific latent expression vector $\mathbf{z}$, combined with the desired view vector $\mathbf{v}$ for rendering.  The output is a collection of 3D primitives $\{\mathcal{V}_k\}$, which cover the occupied regions of the scene; each primitive containing volumetric RGB$\alpha$ information that can be used in traditional volumetric rendering.  Formally,
\begin{equation}
    \{\mathcal{V}_k\} = \text{MVP}(\mathbf{z},\mathbf{v}),
\end{equation}
where each of the $N^2$ primitives is defined by
\begin{equation}
    \mathcal{V}_k = (\mathbf{t}_k, \mathbf{R}_k, \mathbf{s}_k, \mathbf{C}_k).
\end{equation}
Here, the primitive geometry is defined by a translation $\mathbf{t}_k \in \mathbb{R}^3$, a rotation $\mathbf{R}_k \in \text{SO}(3)$, and a non-uniform scale $\mathbf{s}_k \in \mathbb{R}^3$.  The appearance is defined by a dense voxel grid of color information $\mathbf{C}_k \in \mathbb{R}^{4 \times M_x \times M_y \times M_z}$, which stores the RGB$\alpha$ value per voxel (in our implementation, $N^2=4096$ and $M_x = M_y = M_z = M = 16$).

The MVP decoder consists of a geometry branch that depends only on $\mathbf{z}$ and an appearance branch that depends on both $\mathbf{z}$ and $\mathbf{v}$.  
The primitives are organized in a 2D grid of size $N \times N$ and are associated to positions on a guide mesh through a UV parametrization.  The guide mesh is predicted by the geometry branch as a means to initialize the per-primitive transformations, which are further refined in the geometry branch.  
The convolutional appearance branch predicts $\mathbf{C}_k$ directly in the UV-space of the mesh.  We refer to the original formulation~\cite{lombardi2021mvp} for more details.  In our work, we use the MVP geometry branch directly, but we extend the appearance branch to provide the ability to relight the dynamic neural head, as described in \secref{subsec:relightableMVP}.

\subsection{Data Acquisition and Preprocessing}
\label{subsec:data}

Before describing our architecture, we first describe important differences in the training data as compared to the original MVP formulation, which was trained on multi-view video sequences of a performing actor, observed by $\approx$100 different cameras under constant full-on illumination.

\begin{figure}[t]
    \centering
    \includegraphics[width=\columnwidth]{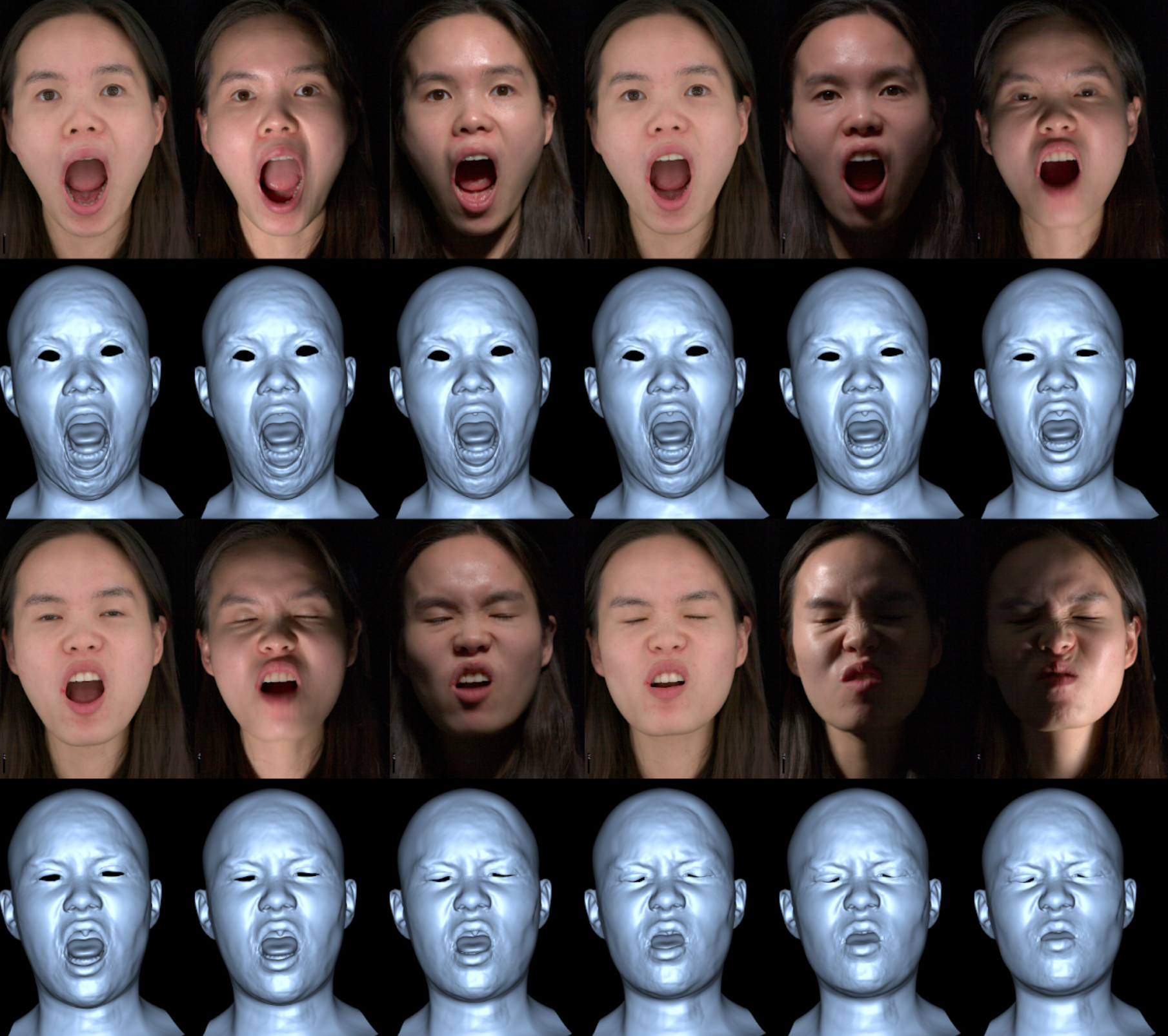}
    \caption{An overview of our training data, which includes dynamic performances illuminated by interleaved OLAT and full-on lighting conditions.  We also obtain per-frame 3D geometry for the face as a representation of the expression.}
    \label{fig:data}
\end{figure}

\paragraph{Image Data.}
Our model is also actor-specific, requiring multi-view video data of a performing subject.  However, as we aim to relight the performances, we require a diverse set of lighting conditions in the dataset.  So instead of constant illumination we capture our subject under a time-varying light pattern consisting of both one-light-at-a-time (OLAT) frames and full-on illumination frames.  In contrast to OLAT acquisition methods that require a dense light stage \cite{debevec2000acquiring,sun2019single,sun2020light,zhu2023neuralrelighting}, we use the camera and light setup of ReNeRF~\cite{xu2023renerf}, which showed impressive relighting of {\em static} neural heads with far less expensive hardware.  The setup consists of only 10 cameras and 32 individually-controllable LED light bars placed in the frontal hemisphere of the capture volume with calibrated 3D positions relative to the cameras~\cite{xu2022eg}.  We propose to capture dynamic performances while flashing a dedicated lighting sequence consisting of one illuminated light bar per frame (OLAT frames).  We also intermix a full-on shot with all light bars illuminated once every 3 frames to help with mesh tracking (as described below).  So the lighting sequence over time is $F, O_1, O_2, F, O_3, O_4, F, ..., F, O_{31}, O_{32}$, where $F$ corresponds to a full-on frame and $O_i$ corresponds to the $i$-th OLAT bar.  The pattern is repeated indefinitely at 24 frames per second.  Importantly, we choose an OLAT ordering such that neighboring OLAT frames are as dissimilar as possible (e.g. an OLAT from the left or top followed by an OLAT from the right or bottom, respectively).  The motivation is that neighboring frames contain very similar facial expressions and so we maximize data efficiency if lighting conditions are as different as possible from one frame to the next.  We do not impose any particular facial expression sequence, but in practice we obtained good results by capturing two different facial muscle workouts followed by three scripted lines of dialog and one free dialog performance.  In total, we capture on average about 2700 images per camera for a single subject.  A short clip from one subject is illustrated in \figref{fig:data}, which shows a partial sequence of OLAT and full-on images from one frontal camera.

\paragraph{Mesh Tracking.}
In addition to the image data, we require a representation of the per-frame facial expressions for training.  To this end, we apply the common practice of pre-computing a tracked 3D mesh sequence corresponding to our input imagery. We employ a recent landmark-based 3D face tracking method~\cite{chandran2023infinitelandmarks}, which optimizes for the parameters of an actor-specific local blendshape model to match detected 2D landmarks in all camera views.  We build the local blendshape model from a small set of 3D face scans in a pre-process.  Tracking is performed only on the full-on illumination frames, and then the model parameters are interpolated linearly across the OLAT frames.  The parameters include both the expression blendweights and the head pose. As we wish to train our network in a stabilized space with respect to the skull position, we use only the expression parameters to construct the face meshes and we use the small per-frame head pose transformations to inversely offset the per-frame camera positions.  The tracked mesh geometry corresponding to a small input sequence is shown alongside the input images in \figref{fig:data}.

To summarize the dataset, the ultimate training data is approximately 1800 frames per capture subject on average (after removing the full-on frames), each frame containing:
\begin{itemize}
    \item 10 multi-view images from calibrated camera positions,
    \item one 3D light position corresponding to the center of the OLAT bar illuminating that frame,
    \item and a tracked 3D face mesh.
\end{itemize}

\subsection{Relightable MVP}
\label{subsec:relightableMVP}

Our new architecture for creating animatable and relightable neural heads can be considered as a relightable version of MVP~\cite{lombardi2021mvp}.  As shown in \figref{fig:overview}, there are three main trainable components (shown in gray): a mesh encoder network to project the input facial expression mesh to a latent global expression parameter $\mathbf{z}$, and parallel geometry and appearance branches similar to MVP as described above.
A small 1-layer MLP designed to prepare $\mathbf{z}$ for the downstream branches is also learned. It maps the 256-dimensional vector $\mathbf{z}$ to a $16384$-dimensional vector that is then reshaped to an $8 \times 8$ feature map with $256$ channels and sent to the convolutional geometry and appearance branches.
For the geometry branch we use exactly the MVP architecture, but we make important changes to the appearance branch to support relighting.  The mesh encoder and the illumination-modulated appearance branches are described in the following.

\paragraph{Mesh Encoder.}
Similar to MVP, we require a latent expression vector $\mathbf{z}$ to drive our neural head decoder.  When tracked geometry is available, the original MVP formulation used the encoder architecture of a Deep Appearance Model~\cite{lombardi2018dam}, which takes the tracked geometry and a color texture as input.  In contrast to MVP, however, we do not bake appearance information into $\mathbf{z}$ and instead drive our model purely from expressions, in the form of tracked geometry alone.  This will allow us to artistically control the neural head at inference time with novel unseen expressions.  We therefore construct our mesh encoder from a modified Deep Appearance Model encoder, specifically omitting the texture branch.  Our encoder is trained end-to-end along with the geometry and appearance decoders.

\paragraph{Illumination-Modulated Appearance.}
The most significant contribution of our work is the proposed illumination-modulated appearance branch.  Rather than conditioning the appearance only on the view vector as in MVP, we additionally consider the per-frame OLAT lighting condition during training.  Notably, we compute per-primitive {\em local} light directions by first evaluating the geometry branch to get the world-space transformations for each primitive, and then computing the grid of local light directions $\mathbf{l}_k$ as the difference between the 3D OLAT light position $\mathbf{p}_{olat}$ and the center of each transformed primitive.  Specifically,
\begin{equation}
    \mathbf{l}_k = \mathbf{p}_{olat} - \mathbf{R}_k \cdot \mathbf{t}_k.
\end{equation}

Conditioning the appearance branch on per-primitive light directions rather than a single global {\em distant} light allows our model to support relighting with nearfield illumination, as we demonstrate in \secref{sec:experiments}.  Analogously, the camera view for rendering is also at a discrete 3D location in the scene, which we denote as $\mathbf{p}_{cam}$, and so we can similarly compute per-primitive {\em local} view directions $\mathbf{v}_k$ rather than a single global view vector $\mathbf{v}$ for conditioning the appearance branch.  Specifically,
\begin{equation}
    \mathbf{v}_k = \mathbf{p}_{cam} - \mathbf{R}_k \cdot \mathbf{t}_k.
\end{equation}

Just as local light directions enable rendering with nearfield illumination, local view directions enable rendering with nearfield camera views, which we demonstrate by synthesizing a dolly-zoom effect in \secref{sec:experiments}.  Both nearfield illumination and nearfield viewpoints are not possible with the original MVP architecture.

Our proposed appearance branch generates the voxel color grids as
\begin{equation}
    \{\mathbf{C}_k\} = \text{RelMVP}(\mathbf{z},\{\mathbf{v}_k\}, \{\mathbf{l}_k\}),
\end{equation}
where we denote RelMVP() as our relightable version of the MVP appearance branch.

The relightable appearance branch is implemented as a convolutional architecture, which takes the reshaped expression vector as input and gradually produces the primitives' RGB$\alpha$ tiles in UV space, modulated by the local view and light directions.  The local view and light directions per primitive are combined and stored as a single 6-channel image in UV space at the full network output resolution (view and light vectors are copied for every voxel within a primitive).  We denote $\mathbf{I} \in \mathbb{R}^{6 \times (N \cdot M) \times (N \cdot M)}$ as the concatenated set of $\{\mathbf{v}_k\}$ and $\{\mathbf{l}_k\}$.
At each convolutional level, $\mathbf{I}$ is bilinearly downsampled and concatenated to the intermediate feature layers before proceeding to the next layer.  This downsampling operation has the effect of averaging local view and light directions across neighboring primitives, which is acceptable since neighboring primitives are located close to each other in 3D space.  At early layers, the averaged view and light directions resemble global view and light vectors, but then at deeper layers the per-primitive view and light directions can specialize the appearance of each primitive individually, allowing us to achieve nearfield lighting and viewpoints.  Note that the local view and light conditioning is only applied to the RGB component of the output, as opacity $\alpha$ is independent of illumination and view direction.

The result is a set of primitive volumes $\{\mathbf{C}_k\}$ that are transformed by the output of the geometry branch and rendered with a differentiable raytracer~\cite{lombardi2021mvp}.

\subsection{Implementation Details}
\label{subsec:implementation}

We employ a fully-convolutional network for the appearance branch RelMVP().
The input is the local light and view vectors, reshaped to a feature map $\mathbf{I} \in \mathbb{R}^{6 \times (N \cdot M) \times (N \cdot M)}$, as well as the expression code $\mathbf{z}$ transformed and reshaped to a feature map $\mathbf{z}' \in \mathbb{R}^{256 \times 8 \times 8}$ (channels $\times$ height $\times$ width).
The appearance branch then consists of seven transpose-convolution layers with a kernel size of 4, stride 2 and padding 1, which increase the feature map resolution from $8 \times 8$ by a factor of two at every step until the final resolution of $1024 \times 1024$ (i.e. $N \cdot M = 1024$) is reached.
The inputs to the convolutional layers are the previous feature maps, starting with $\mathbf{z}'$ and the six channels from $\mathbf{I}$ bilinearly downsampled to match the current spatial resolution. The output features have channels $256, 128, 128, 64, 64, 32, 48$ where the final 48 channels are interpreted as $rgb \times M_z$.

Since opacity $\alpha$ does not depend on the local view or light direction, we follow the pracitce of the original MVP architecture~\cite{lombardi2021mvp} and estimate opacity in a separate branch. This branch is identical to the above architecture, with the difference of predicting $M_z$ output channels and not depending on $\mathbf{I}$. We apply ReLU activation on the RGB output and all intermediate layers are followed by LeakyReLU activations.

Our neural rendering pipeline is trained end-to-end on multi-view OLAT sequences, see \secref{subsec:data}. 
We drop the background model that estimates foreground and background objects in the original MVP architecture since our data is recorded in front of a pure black backdrop.
Instead we add a matting loss to the MVP loss functions that compares a target matting mask $\mathcal{M}$ to the accumulated density per ray $\tilde{\alpha}(\Theta)$, which avoids floating primitives at the background,
\begin{equation}
    \mathcal{L}_\text{mat} := \text{MAE}(\mathcal{M}, \tilde{\alpha}(\Theta)) .
\end{equation}
We extract the matting masks $\mathcal{M}$ from the input images using MODNet~\cite{Ke2022modnet}.
The final loss function is then
\begin{equation}
    \begin{aligned}
        \mathcal{L} :=& 
            \lambda_\text{pho} \mathcal{L}_\text{pho} + 
            \lambda_\text{geo} \mathcal{L}_\text{geo} + 
            \lambda_\text{vol} \mathcal{L}_\text{vol} \\ +& 
            \lambda_\text{kld} \mathcal{L}_\text{kld} + 
            \lambda_\text{mat} \mathcal{L}_\text{mat}
    \end{aligned}
\end{equation}
where all loss terms other than $\mathcal{L}_\text{mat}$ are the same as in the MVP implementation, and the weights are defined as
\begin{equation}
    \begin{aligned}
    &\lambda_\text{pho}=1.0,\  
    \lambda_\text{geo}=0.1,\  
    \lambda_\text{vol}=0.01,\\  
    &\lambda_\text{kld}=0.001,\  
    \lambda_\text{mat}=0.1 .\ 
\end{aligned}
\end{equation}
We employ ADAM~\cite{Kingma2014Adam} as the optimizer with a learning rate of $lr=0.0001$.
For training and evaluation, we downsample the input images to a resolution of $1024 \times 768$ to reduce the training time.
Each subject is trained for 200,000 iterations with a batch size of 12, taking around two days to train on an A6000 GPU.

%% file: sec/4_experiments.tex
\section{Experiments}
\label{sec:experiments}

In this section we perform several experiments to validate our method.  We start with a number of qualitative results, including relighting our dynamic neural heads with novel viewpoints, expressions and illumination conditions (\secref{subsec:qualitative}).  We then perform a quantitative evaluation to show the performance of our model on held-out validation data, including a comparison to related work (\secref{subsec:quantitative}). For all results, we recommend to also view the accompanying supplemental video, in order to see the animations in motion.

\subsection{Qualitative Results}
\label{subsec:qualitative}

\begin{figure}[t]
    \centering
    \includegraphics[width=\columnwidth]{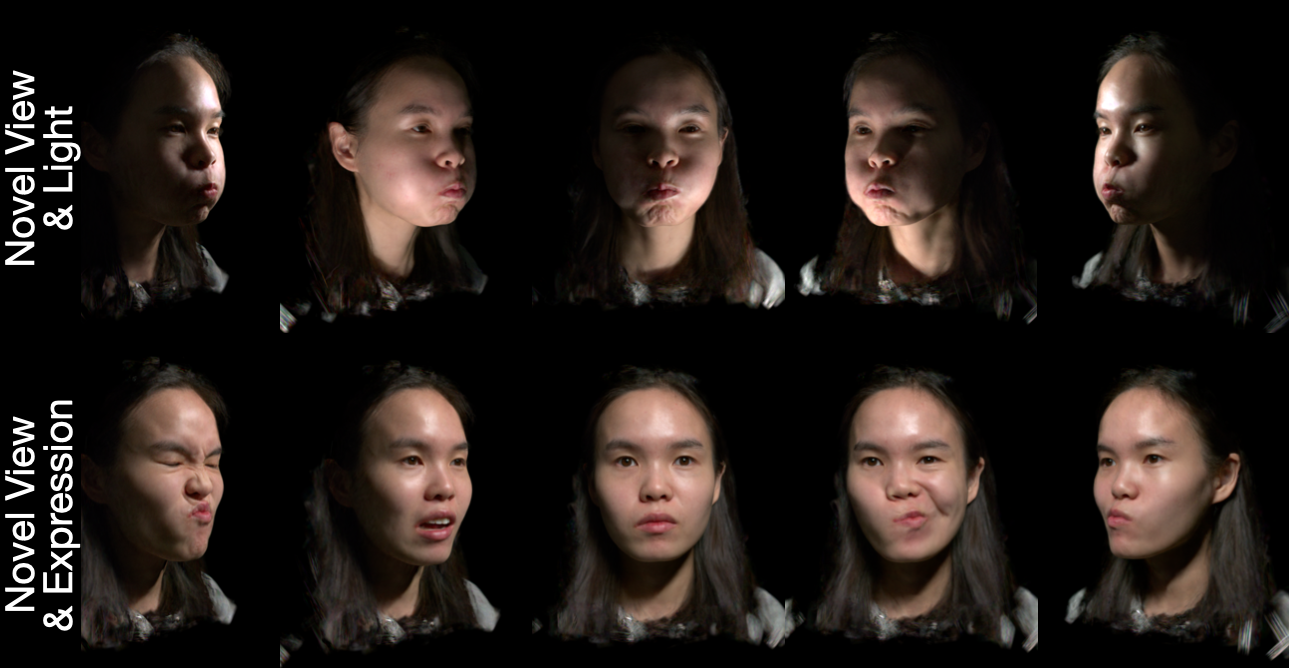}
    \caption{Novel view, light, and expressions: Our method allows for artistically-generated renders under novel point lights (top) and novel expressions (bottom), all rendered from novel viewpoints.}
    \label{fig:novelViewLightExpStudio}
\end{figure}

We begin by showing the ability of our model to re-render our captured neural heads from novel viewpoints with artist-controllable lighting and expressions.  \figref{fig:novelViewLightExpStudio} depicts one of our subjects in the studio capture environment, but relit with novel point lights (top row) and novel interpolated expressions (bottom row).  All renders are generated from unseen viewpoints outside of the training views.

\begin{figure}[t]
    \centering
    \includegraphics[width=\columnwidth]{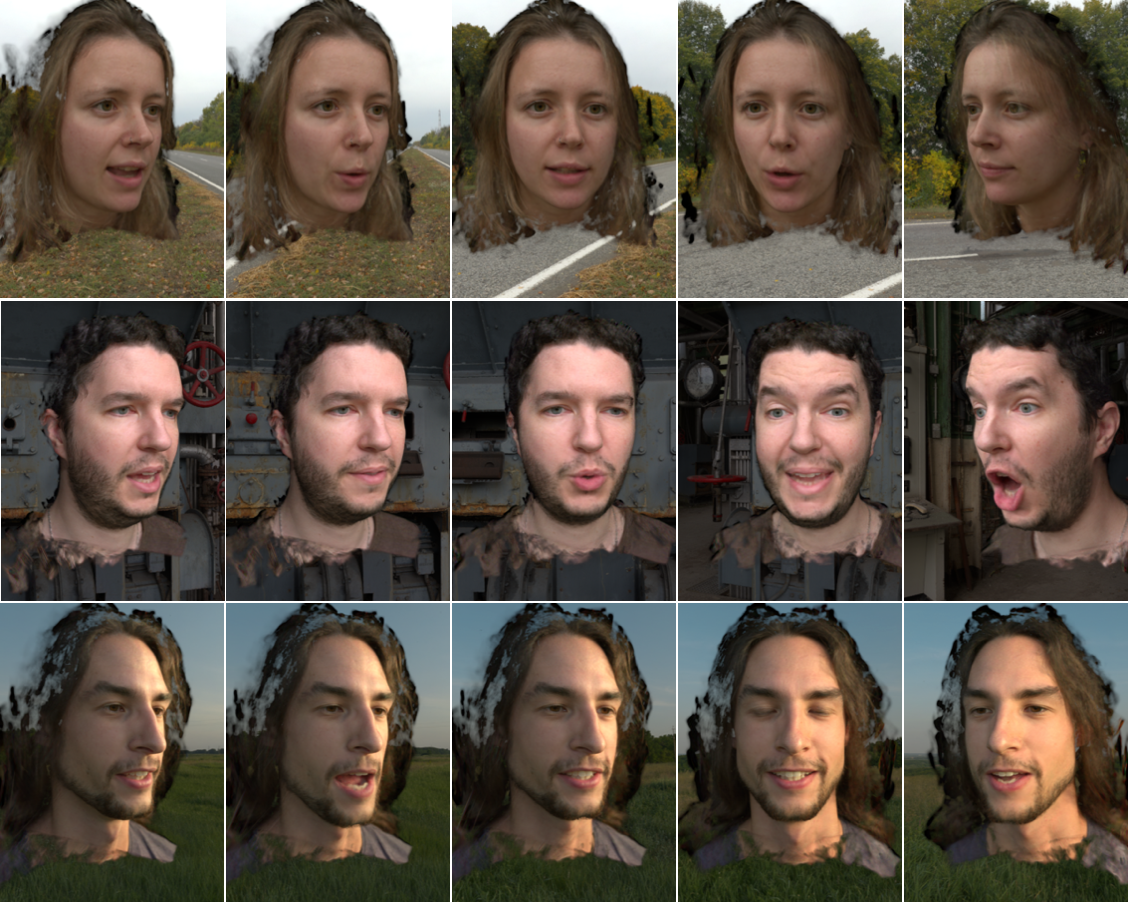}
    \caption{Novel environment maps: Since our method generalizes to novel light directions, we can densely sample light directions over the hemisphere to render performances under any environment map.  Three examples are shown, rendered from novel viewpoints.}
    \label{fig:envMapNovelViews}
\end{figure}

In \figref{fig:envMapNovelViews} we highlight the application of re-rendering dynamic performances under arbitrary environment map illumination.  Here we also show the performances under a turntable of novel viewpoints.  Even though our training data consisted only of OLAT lighting frames, we can sample multiple individual light directions from environment maps and combine the rendered results into final high quality and consistent renders.

\begin{figure}[t]
    \centering
    \includegraphics[width=\columnwidth]{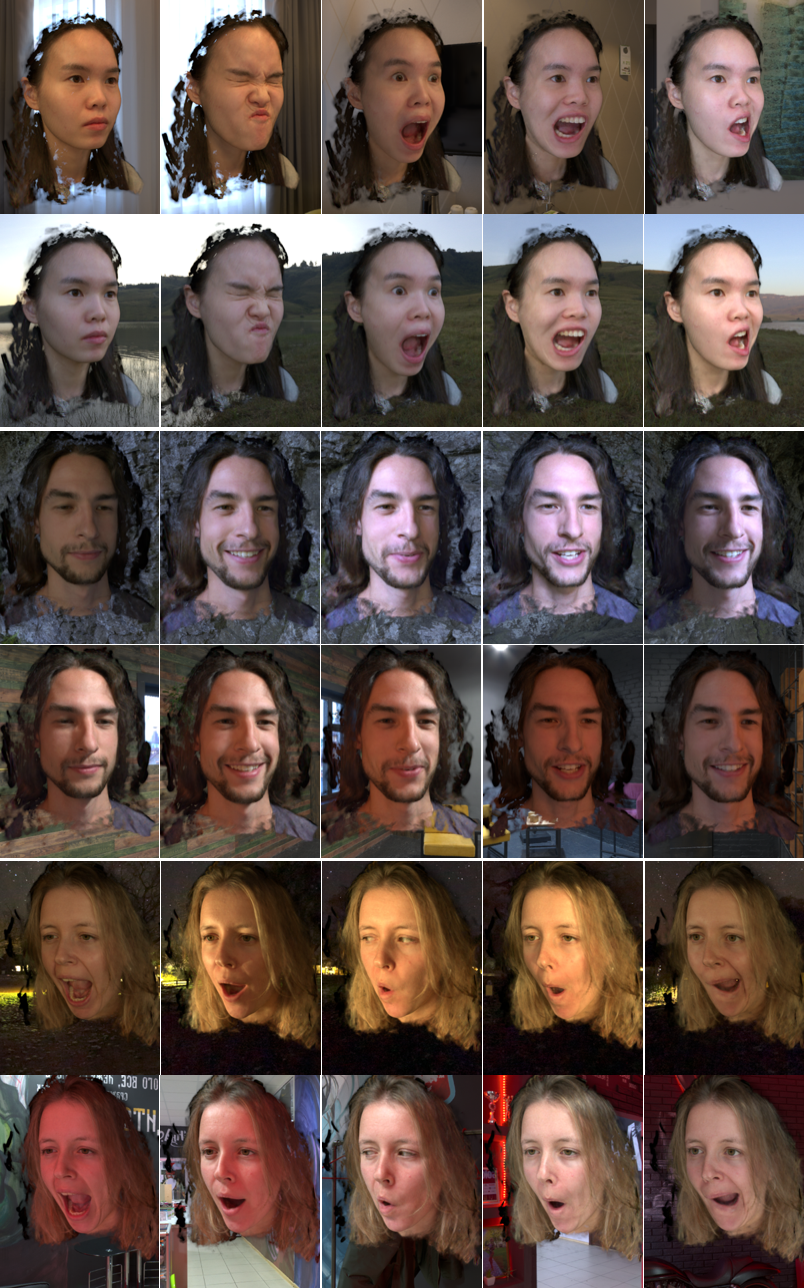}
    \caption{Here we demonstrate the quality our method achieves on further subjects: we can visualize performances under novel views and arbitrary, temporally rotated environment maps that produce complex lighting conditions.}
    \label{fig:heroEnvMapRotate}
\vspace{-6mm}
\end{figure}

We further push the capabilities of our model by rendering several performances under a number of challenging light conditions in \figref{fig:heroEnvMapRotate}.  Here we use six different environment maps from both indoor and outdoor scenes, during daytime and night, and we render the dynamic neural heads of 3 subjects from a fixed camera view while rotating the environment maps around the subjects.  The two rows belonging to the same subject show the same expression but rendered with different lighting conditions, highlighting the versatility of our method for relighting animated performances.

\begin{figure}[t]
    \centering
    \includegraphics[width=\columnwidth]{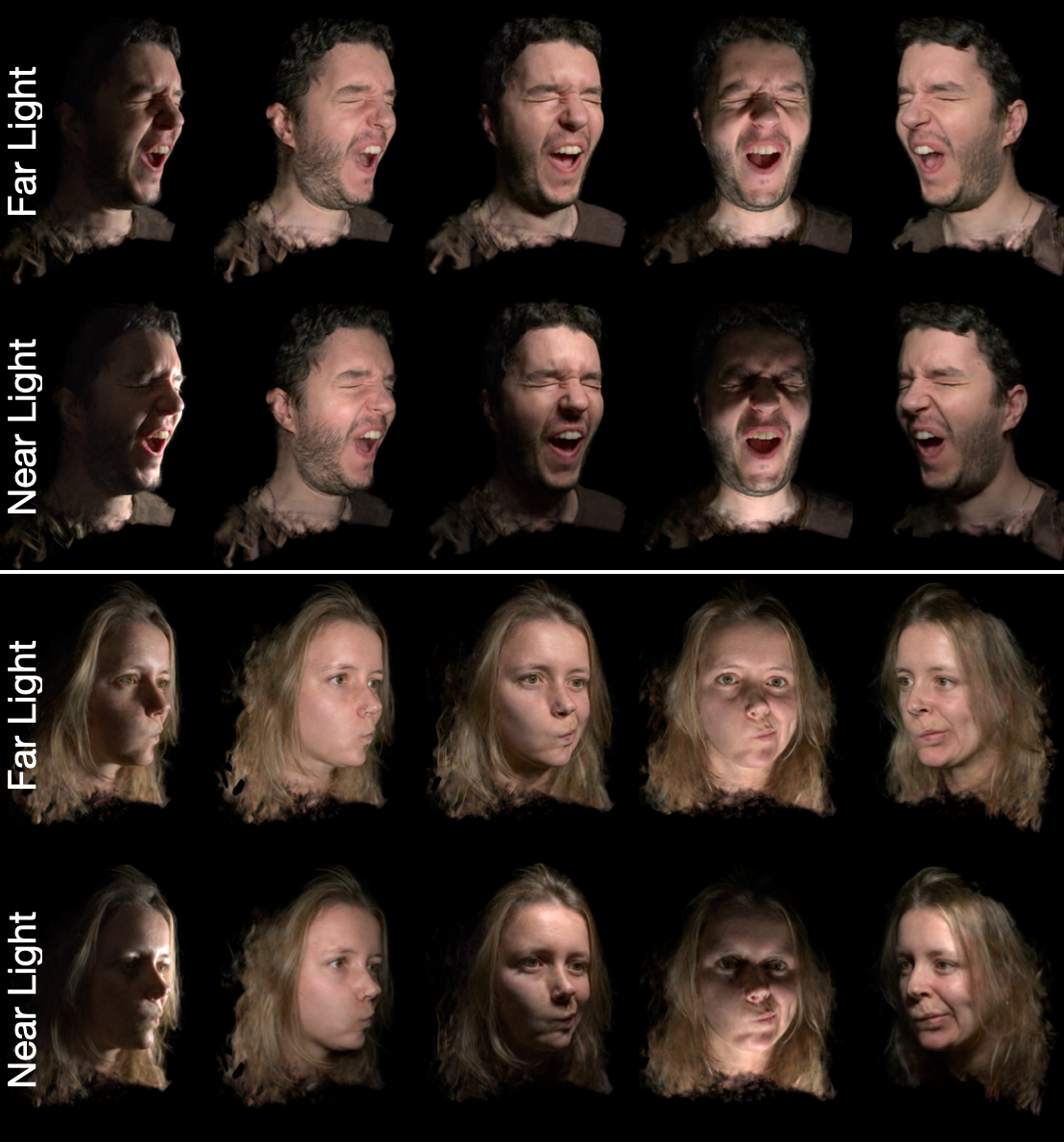}
    \caption{Near-field lighting: By specifying a per-primitive light direction in the appearance branch, we can render both far-field lights, as well as near-field lights.}
    \label{fig:nearfieldLighting}
\end{figure}

\begin{figure}[t]
    \centering
    \includegraphics[width=\columnwidth]{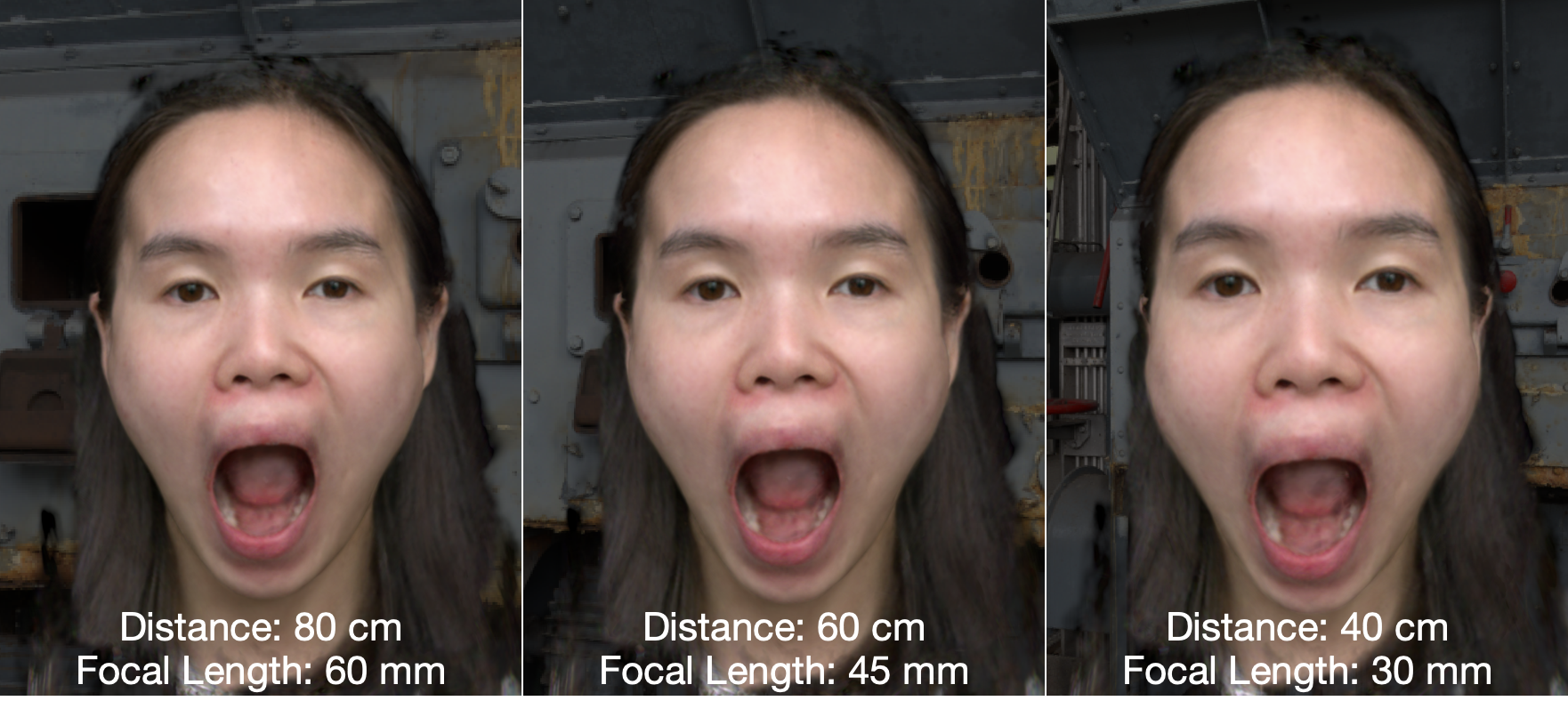}
    \caption{Near-field view: By conditioning the appearance branch also on per-primitive view directions, we can change the focal length of the camera and realize effects like a dolly zoom.}
    \label{fig:nearfieldView}
\end{figure}

An important aspect of our architecture is that the light and view directions are computed locally per primitive, allowing nearfield lighting and viewpoint effects.  We demonstrate nearfield lighting in \figref{fig:nearfieldLighting}, which shows a static expression of 2 captured subjects relit by a moving point light source (and varying unseen viewpoints).  In the top row of each subject the point light is farther away from the subject than the corresponding frame in the bottom row, where the light is near to the face.  Our method supports natural relighting under these conditions.  In a similar spirit, having per-primitive local view directions allows us to synthesize complex camera motions like a dolly-zoom effect, where the camera pushes in close on a subject's face while decreasing the focal length of the lens (i.e. increasing the FOV).  We demonstrate this effect in \figref{fig:nearfieldView}, which shows a realistic simulation of this commonly-used practical camera move.

The presented results show a non-exhaustive sample set of applications that are enabled by our artist-friendly method for relightable and animatable neural heads.

\subsection{Quantitative Evaluation}
\label{subsec:quantitative}

To evaluate our method quantitatively we trained a version with some held-out data for validation.  Specifically, we constructed three validation sets: one with held-out OLAT directions, one with held-out performances, and one combined one that contains both held-out light directions and performances.  Our validation data consists of a variety of frames from three different captured subjects.

As we evaluate our model's performance, we simultaneously perform a comparison to related work.  Unfortunately there are very few existing methods for both relightable and animatable neural avatars, in particular with code available for testing. 
To conduct a comparison with a baseline method, we re-implemented the lighting branch of TRAvatar~\cite{yang2023towards}, denoted by TRAvatar* below.

\begin{figure}[t]
	\centering
	\includegraphics[width=\columnwidth]{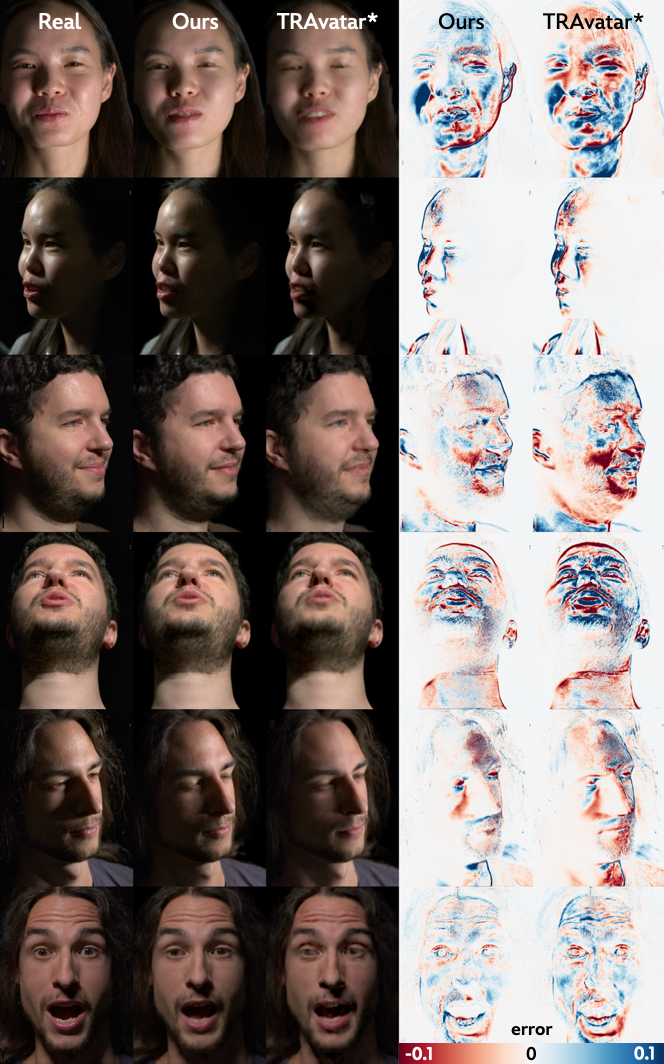}
	\caption{Example frames of the quantitative comparison against TRAvatar* on held-out data, see \secref{subsec:quantitative}. The comparisons were conducted on three subjects, called S1 to S3 from top to bottom. The heatmaps to the right depict the per-pixel errors of our method and TRAvatar* against the ground truth images.}
	\label{fig:comparison}
\end{figure}

\begin{table}
    \caption{Quantitative evaluation of our method against TRAvatar*. Results are shown for 3 subjects (see \figref{fig:comparison}), evaluated for novel light directions, novel performances, and both novel light directions and performances. Our method consistently outperforms TRAvatar.}
    \label{tab:quantComparison}
    \footnotesize
    \begin{tabular}{rrr|cccc}
        \toprule
        &&& PSNR $\uparrow$ & MAE $\downarrow$ & SSIM $\downarrow$ & LPIPS $\downarrow$ \\

        \midrule
        \parbox[t]{2mm}{\multirow{6}{*}{\rotatebox[origin=c]{90}{novel light}}} &
        \parbox[t]{2mm}{\multirow{2}{*}{\rotatebox[origin=c]{90}{S 1}}} &
           ours & \textbf{32.19} & \textbf{2.88} & \textbf{0.899} & \textbf{0.278} \\
        && TRAvatar* & 29.68 & 4.03 & 0.870 & 0.326 \\
        \cmidrule(r){2-7}
        & \parbox[t]{2mm}{\multirow{2}{*}{\rotatebox[origin=c]{90}{S 2}}} &
           ours & \textbf{32.20} & \textbf{3.11} & \textbf{0.864} & \textbf{0.296} \\
        && TRAvatar* & 30.13 & 4.06 & 0.832 & 0.355 \\
        \cmidrule(r){2-7}
        & \parbox[t]{2mm}{\multirow{2}{*}{\rotatebox[origin=c]{90}{S 3}}} &
           ours & \textbf{33.73} & \textbf{2.65} & \textbf{0.873} & \textbf{0.343} \\
        && TRAvatar* & 32.37 & 3.24 & 0.854 & 0.390 \\

        \midrule
        \parbox[t]{2mm}{\multirow{6}{*}{\rotatebox[origin=c]{90}{novel perf.}}} &
        \parbox[t]{2mm}{\multirow{2}{*}{\rotatebox[origin=c]{90}{S 1}}} &
           ours & \textbf{39.75} & \textbf{3.70} & \textbf{0.881} & \textbf{0.282} \\
        && TRAvatar* & 28.86 & 4.12 & 0.869 & 0.323 \\
        \cmidrule(r){2-7}
        & \parbox[t]{2mm}{\multirow{2}{*}{\rotatebox[origin=c]{90}{S 2}}} &
           ours & \textbf{29.52} & \textbf{4.09} & \textbf{0.835} & \textbf{0.300} \\
        && TRAvatar* & 28.88 & 4.46 & 0.829 & 0.350 \\
        \cmidrule(r){2-7}
        & \parbox[t]{2mm}{\multirow{2}{*}{\rotatebox[origin=c]{90}{S 3}}} &
           ours & \textbf{31.28} & \textbf{3.27} & \textbf{0.863} & \textbf{0.338} \\
        && TRAvatar* & 31.21 & 3.36 & 0.861 & 0.373 \\

        \midrule
        \parbox[t]{2mm}{\multirow{6}{*}{\rotatebox[origin=c]{90}{light \& perf.}}} &
        \parbox[t]{2mm}{\multirow{2}{*}{\rotatebox[origin=c]{90}{S 1}}} &
           ours & \textbf{28.68} & \textbf{4.30} & \textbf{0.865} & \textbf{0.300} \\
        && TRAvatar* & 27.30 & 5.27 & 0.844 & 0.342 \\
        \cmidrule(r){2-7}
        & \parbox[t]{2mm}{\multirow{2}{*}{\rotatebox[origin=c]{90}{S 2}}} &
           ours & \textbf{29.11} & \textbf{4.34} & \textbf{0.826} & \textbf{0.316} \\
        && TRAvatar* & 28.13 & 4.94 & 0.812 & 0.366\\
        \cmidrule(r){2-7}
        & \parbox[t]{2mm}{\multirow{2}{*}{\rotatebox[origin=c]{90}{S 3}}} &
           ours & \textbf{30.76} & \textbf{3.55} & \textbf{0.855} & \textbf{0.354} \\
        && TRAvatar* & 30.65 & 3.80 & 0.848 & 0.392 \\

        \bottomrule
    \end{tabular}
\end{table}

\tabref{tab:quantComparison} shows the comparisons between our method and TRAvatar* on the held-out data. Example frames from the comparison with a heatmap visualizing the per-pixel errors are shown in \figref{fig:comparison}. Note that given a novel directional light, TRAvatar can only evaluate it using barycentric coordinates within their fixed basis used in training.
Our method clearly outperforms TRAvatar* in terms of both qualitative render quality and quantitative analysis in all cases.  Furthermore, as mentioned earlier, TRAvatar* is unable to interpolate novel lighting directions or model nearfield effects, which our method can easily achieve.

%% file: sec/5_conclusion.tex
\section{Conclusion}
\label{sec:conclusion}

We present a novel architecture for relightable, animatable neural avatars,
building on top of the Mixture of Volumetric Primitives architecture.
Our architecture extends the appearance branch with per-primitive light and view directions, allowing for nearfield lighting and viewpoint effects. The networks are trained end-to-end on OLAT sequences obtained by flashing 32 LED bars during a dynamic performance of a subject, captured with inexpensive hardware.
Our model is capable of novel-view and novel-light estimation and also generalizes well to novel expressions and performances.

We note a few practical limitations of our method.  So far, we have focused primarily on the frontal part of the head and do not recover a complete 360-degree neural avatar.  This can result in artifacts at the boundary (\eg, the hair and neck regions).  That said, we do not believe that our model is fundamentally restricted here, and our results are only limited by the physical capture setup we used.  Further, during very fast motions, if motion-blur were to occur in the training data then this will lead to blurry reconstructions.  We also found that extrapolation to extreme facial expressions outside our training data was not possible, but we note that our results were generated from an order of magnitude fewer training frames than the original MVP algorithm, and therefore simply adding additional expression variation during training would help alleviate this problem. Finally, as our input to the network is a mesh, gaze changes and hair motions are currently not controllable.

%% file: sec/X_suppl.tex
\clearpage
\setcounter{page}{1}
\maketitlesupplementary

In this supplemental document we discuss additional qualitative results, experimental details and ablations in \secref{sec:supp_experiments}, and end with limitations and future work in \secref{sec:supp_futurework}.

\section{Experiments}
\label{sec:supp_experiments}

\subsection{Additional Qualitative Results}

Here we illustrate additional results of novel animation sequences and lighting interpolation \& extrapolation.

\begin{figure}[b]
	\centering
	\includegraphics[width=\columnwidth]{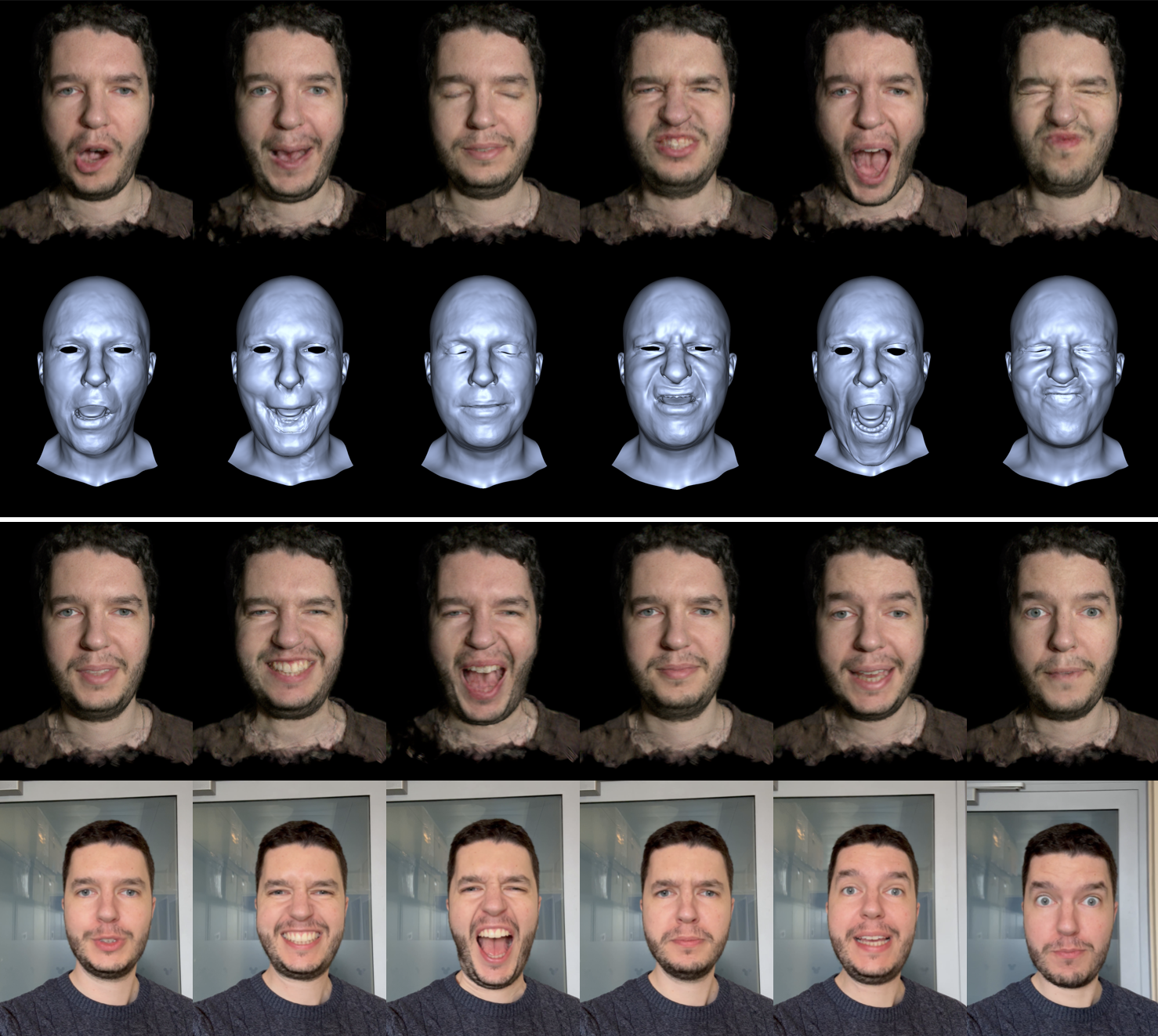}
	\caption{Our method can render novel expressions driven by performances from different sources, \eg artistically-created blendshape animations (top) or monocular face capture from a mobile phone (bottom).}
	\label{fig:novel-perf}
\end{figure}

\paragraph{Novel Animations.} 
As our model takes only 3D meshes as input, it can generate novel controllable animations driven by different sources. Two different examples are shown in \figref{fig:novel-perf}.  In the first example (top 2 rows), an artist-created blendshape animation is used to drive our model.  We show our result alongside the 3D mesh animation.  In the second example (bottom 2 rows), we drive our model from monocular video input.  Here we track the performance of the actor using a recent landmark-based 3D facial reconstruction method~\cite{chandran2023infinitelandmarks}, and then feed the tracked meshes into our model.  We show our result alongside the original video input.  Given these two examples, we can see that our model can generate realistic renderings of various novel expressions outside of the training sequences. 

\begin{figure*}
	\centering
	\includegraphics[width=\textwidth]{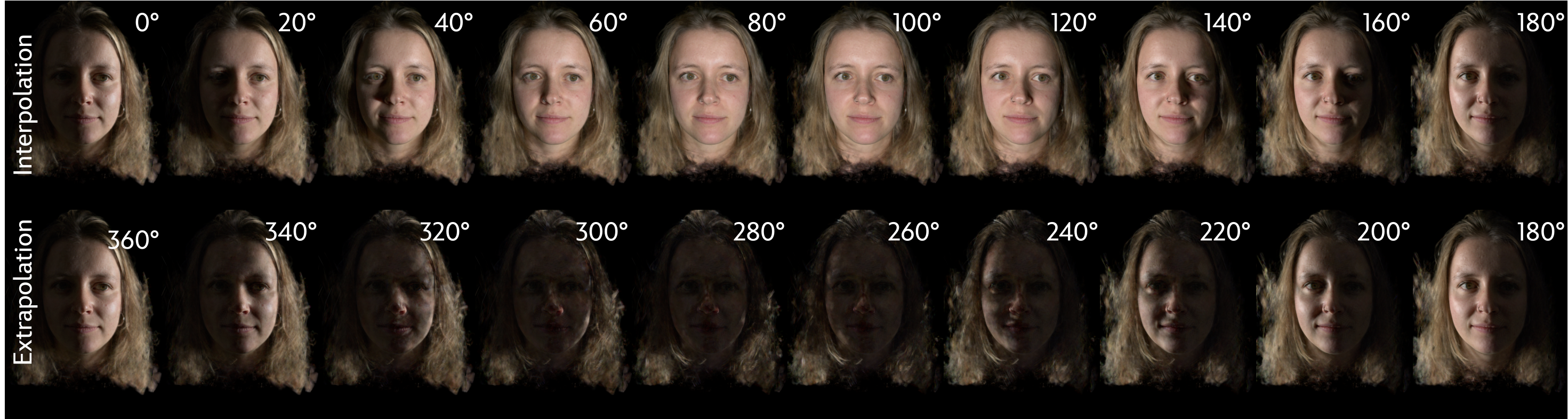}
	\caption{Relighting results of a point light orbiting 360$^\circ$ around the head. The top row (0$^\circ$ to 180$^\circ$) shows our model can interpolate smoothly within the range of the training lighting directions. The bottom row (180$^\circ$ to 360$^\circ$) shows our model can extrapolate to at least 20$^\circ$ on both sides and predict reasonable shadow distributions far outside the training data.}
	\label{fig:interp-extrap}
\end{figure*}

\paragraph{Lighting Interpolation and Extrapolation.} 
To evaluate how the model interpolates and extrapolates novel lighting directions, we render in \figref{fig:interp-extrap} a point light orbiting around the head within a horizontal plane at a radius of 3 meters, starting from the left side of the face. As shown in the figure, our model can smoothly interpolate inside the range of the training lighting directions (0$^{\circ}$ to 180$^{\circ}$), with coherently moving shadows and specular highlights. Although our 32 lights are spread out to cover only the frontal hemisphere, the model can extrapolate well to at least 20$^{\circ}$ towards the backside. While we see extrapolation artifacts beyond that, our model learns to predict reasonable shadow distributions even directly behind the captured subject. 

\subsection{Experiment Details}

Here we provide further details regarding the quantitative evaluation in the main text, and a thorough explanation of how we render images with LatLong environment maps.

\begin{figure}[t]
	\centering
	\includegraphics[width=\columnwidth]{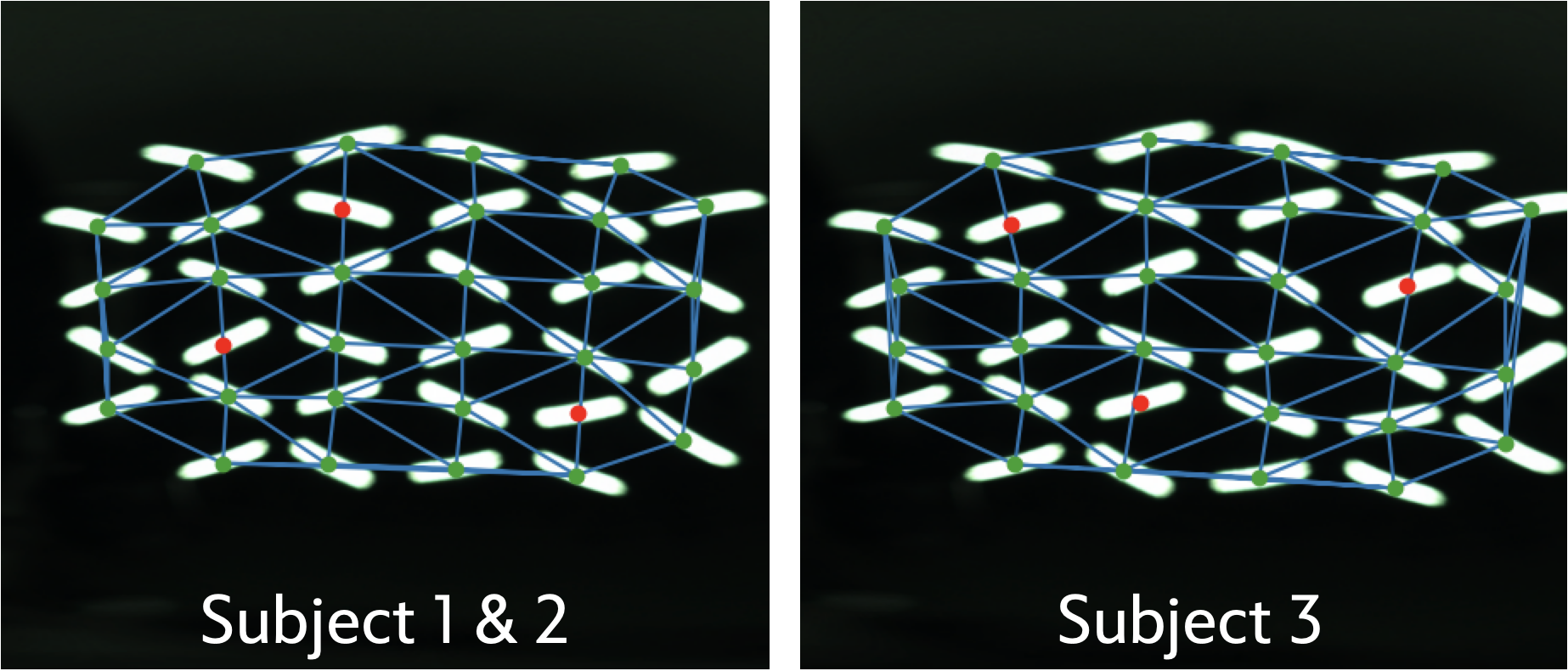}
	\caption{Captured light probe in Latlong format overlaid with computed triangulation of the training lights. Training lights are marked with green dots and held-out lights are marked in red.}
	\label{fig:heldout}
\end{figure}

\paragraph{Quantitative Evaluation.} 
For the quantitative evaluation experiment in \secref{subsec:quantitative}, we leave out 3 lights for each subject. \figref{fig:heldout} shows the LatLong images of the captured light probe, where each LED bar is represented with a single position/direction. Lights used in training are illustrated as green dots and held-out lights are labeled in red. Note that we did not leave out any lights on the boundary as TRAvatar \cite{yang2023towards} cannot extrapolate outside the training lighting directions. We also compute the Delaunay triangulation of the training lights and use barycentric coordinates to evaluate held-out lights for TRAvatar*, as shown in \figref{fig:heldout}. Similar to ReNeRF \cite{xu2023renerf}, we optimize per LED-bar intensity during training. When testing novel lights, we scale the predicted renders to match the scale of the ground truth images for all frames before computing the metrics in \tabref{tab:quantComparison}. We leave out the free dialog performance of Subject 1, and a scripted line of Subject 2 and Subject 3 as the validation set for novel performances. All numbers in \tabref{tab:quantComparison} are computed in linear RGB space.

\paragraph{Rendering with LatLong Environment Maps.} 
When rendering novel environment maps, our model follows the approach of image-based relighting. More specifically, we downsample the LatLong into uniformly distributed directions, render each of these directional lights, then compute a weighted sum in image space, with weights given by the light intensities. We drop the 10\% of the lights with the lowest intensities to speed up rendering. To avoid extrapolation artifacts (\figref{fig:interp-extrap}), we mask out directions that fall into the 45$^\circ$ cone whose axis points to the back. For the rest of the directions in the back hemisphere, we apply an attenuation term $a = \cos^4 \theta$ ($a$ is 1 at the side and 0 at the back) on their intensities to render smooth environment map animations. The effective resolution of all environment maps we use is approximately 256 directions. For more environment relighting results, please refer to the supplementary video.

\subsection{Ablations}
\begin{figure}
	\centering
	\includegraphics[width=\columnwidth]{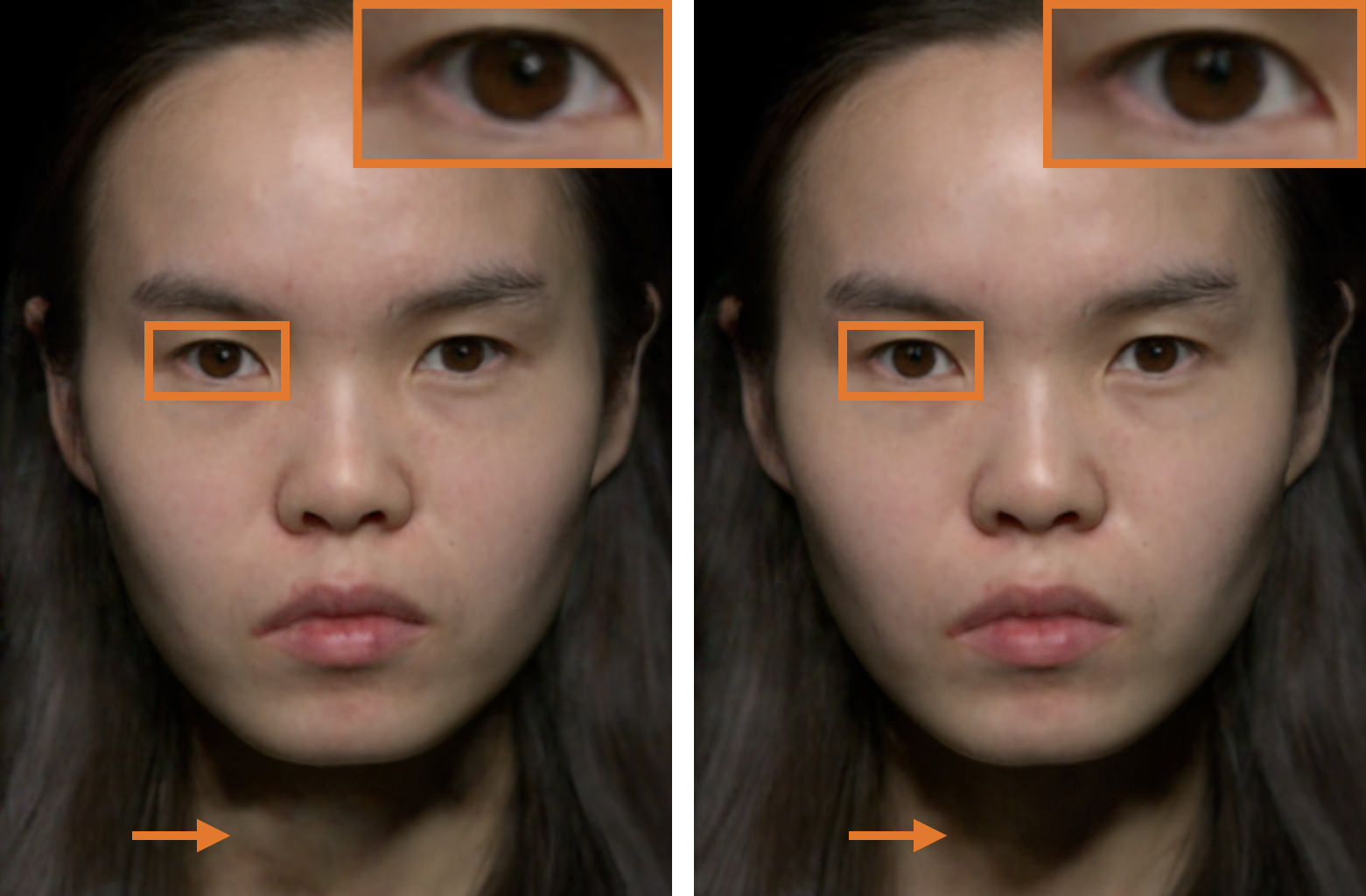}
	\caption{Rendering of a novel point light from our method (left) and our method with learned OLAT codes \cite{xu2023renerf} (right). The learned OLAT codes improve  interpolation of shadows but lead to less sharp rendering of specular reflections on the eyes.}
	\label{fig:codebook}
\end{figure}

In this section, we compare our method with a modified version that incorporates the spherical codebook proposed in ReNeRF \cite{xu2023renerf}. Instead of representing the lights as 3D vectors $\mathbf{l}_k$, we use 64D OLAT codes learned from a small 3-layer MLP. Each layer has 64 hidden units. The input light directions are position-encoded with a 5th-order spherical harmonics basis. While the differences in overall relighting quality across multiple frames are hardly noticeable, incorporating the learned OLAT codes does improve the shadow boundaries in some cases. However, it also impairs the specular reflections on the eyes. We show such an example of a novel point light relighting in \figref{fig:codebook}, where the rendering of our method is on the left, and our method with OLAT codes is on the right.

\section{Limitations \& Future Work}
\label{sec:supp_futurework}

\begin{figure}[t]
	\centering
	\includegraphics[width=\columnwidth]{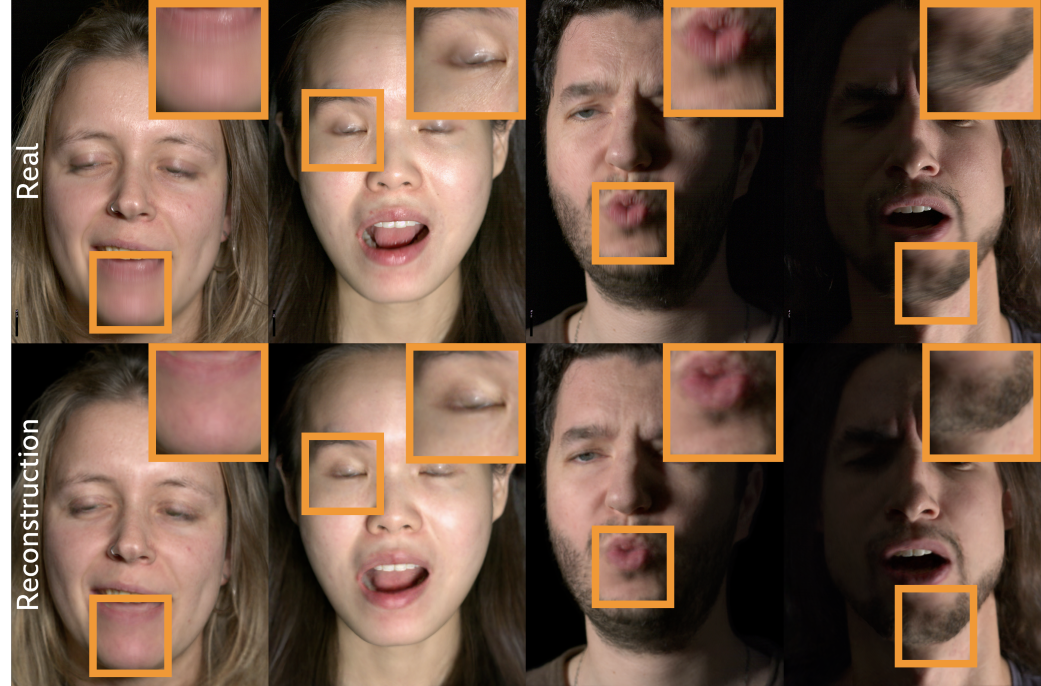}
	\caption{We capture motion blur in the training data due to our low frame rate, which can lead to blurry rendering of fast motions.}
	\label{fig:blur}
\end{figure}

\begin{figure}[t]
	\centering
	\includegraphics[width=\columnwidth]{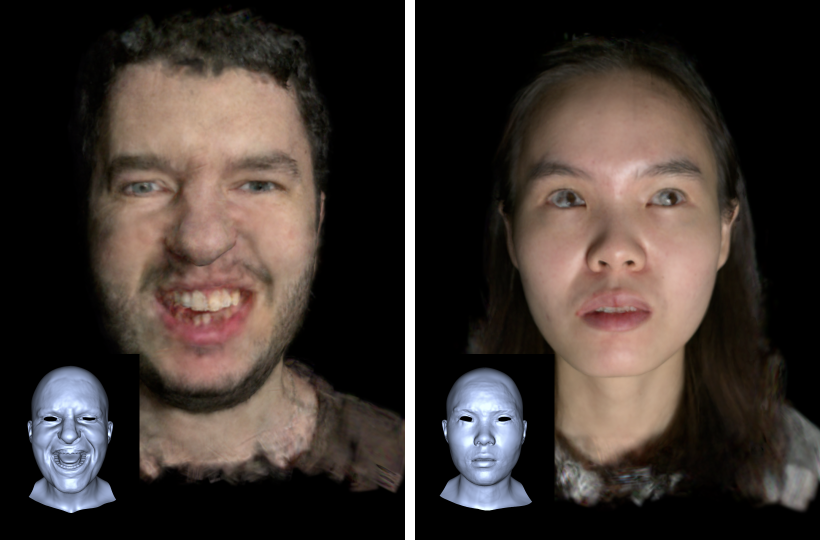}
	\caption{Failure cases. Our model breaks when extrapolating to extreme expressions far from the training data (left). And gazes cannot be disentangled from the base meshes (right). }
	\label{fig:fail}
\end{figure}

One limitation of our method is that it does not achieve real-time performance. It takes about 15s per frame to render an environment map with 256 directions. As mentioned in the main text, we capture motion blur in some of our training frames due to our less expensive setup compared to other methods \cite{bi2021deep, yang2023towards, Meka:2020}. This can lead to blurry rendering for fast motions such as blinking. However, we show in  \figref{fig:blur} that our model does not overfit to all these blurry pixels and is able to recover some details lost in the captured images. Future work could consider modeling motion blur explicitly to render sharper images. Finally, in \figref{fig:fail}, we show failure examples of our method when extrapolating to extreme expressions far from the training data (left), and when the gaze is entangled with the base geometry (right). For future work, we plan to collect more data, which can help with expression extrapolation. Currently, our models are trained with only 10\% of the amount of data as compared to the original MVP algorithm. We also plan to model gazes explicitly to allow for gaze animation.